\documentclass[10pt,twocolumn,letterpaper]{article}

\usepackage[pagenumbers]{cvpr} % arXiv version: authors and page numbers, without review markings
\usepackage{tikz}

\usepackage{tabularx}
\usepackage{colortbl}

\definecolor{EPModelBlue}{HTML}{D0E7FF}
\definecolor{EPExposureRed}{HTML}{FF0000}
\newcommand{\EPmodelrow}{\rowcolor{EPModelBlue}}
\DeclareRobustCommand{\EPstudent}[1]{\textbf{#1}}
\DeclareRobustCommand{\EPresult}[1]{{\bfseries\boldmath #1}}
\DeclareRobustCommand{\EPbest}[1]{\underline{#1}}
\DeclareRobustCommand{\EPseen}{\textcolor{EPExposureRed}{\textit{(seen)}}}
\DeclareRobustCommand{\EPunseen}{\textcolor{EPExposureRed}{\textit{(unseen)}}}
\DeclareRobustCommand{\EPup}{\ensuremath{\,\uparrow}}

\usepackage{stfloats}

\definecolor{cvprblue}{rgb}{0.21,0.49,0.74}
\usepackage[pagebackref,breaklinks,colorlinks,allcolors=cvprblue]{hyperref}

\def\confName{CVPR}
\def\confYear{2027}

\title{EmbeddingProstate: A Unified Model Integrating Diagnostic, Spatial, and Semantic Representations for Prostate MRI}

\author{Zhonghao Liu$^{1}$ \quad Hanxue Gu$^{1}$ \quad Yu-Dong Zhang$^{2}$\\
Jie Bao$^{1,3}$ \quad Kang Wang$^{1}$ \quad Yang Yang$^{1}$\\[4pt]
$^{1}$University of California, San Francisco\\
$^{2}$Nanjing Medical University\\
$^{3}$The First Affiliated Hospital of Soochow University
}

\begin{document}
\maketitle
% Write your text in the two section files below.
% Original CVPR author-kit examples are preserved in template_reference/.
\begin{abstract}
Training examples for prostate MRI are scattered across institutions, tasks, and training objectives. A segmentation model is trained on the exams with lesion masks, a classification model on the exams labeled for clinically significant prostate cancer (csPCa), and an image--report model on the exams with reports, which leaves several specialist models, each optimized on its own data, and a radiologist needs all of them to read one exam. Hence, we investigate whether the same training examples can instead produce one model that matches or even beats each specialist. We present EmbeddingProstate, a unified model for prostate MRI distilled from three specialists, which performs csPCa classification, lesion segmentation, and image--report retrieval with a single encoder. In addition, we introduce Teacher-Residual Supervision, which reduces the uneven overfitting across tasks that occurs when ground-truth labels are added to distillation. Extensive experiments on prostate MRI from three hospitals and two public datasets (PI-CAI and PROMIS) show that EmbeddingProstate matches the specialists in csPCa classification and lesion segmentation, with average gains of 2.5\% in AUROC and 3.4\% in Dice, and improves image-to-report R@10 by 27.4\%. As a foundation model, its frozen encoder also outperforms five public encoders on three external datasets, with an average gain of 18.4\% in linear-probe csPCa AUROC over the best of them.
To our knowledge, EmbeddingProstate is both the first foundation model for prostate MRI and the first model specifically developed for image–report alignment in prostate MRI. We will publicly release the code and pretrained weights upon acceptance.
\end{abstract}

\section{Introduction}
\label{sec:intro}

% Page-one right column, 3.28125 in (83.34 mm) wide.
% Editable Apricot artwork: figure1_three_outputs.pptx.
\begin{figure}[!t]
  \centering
  \includegraphics[width=\columnwidth]{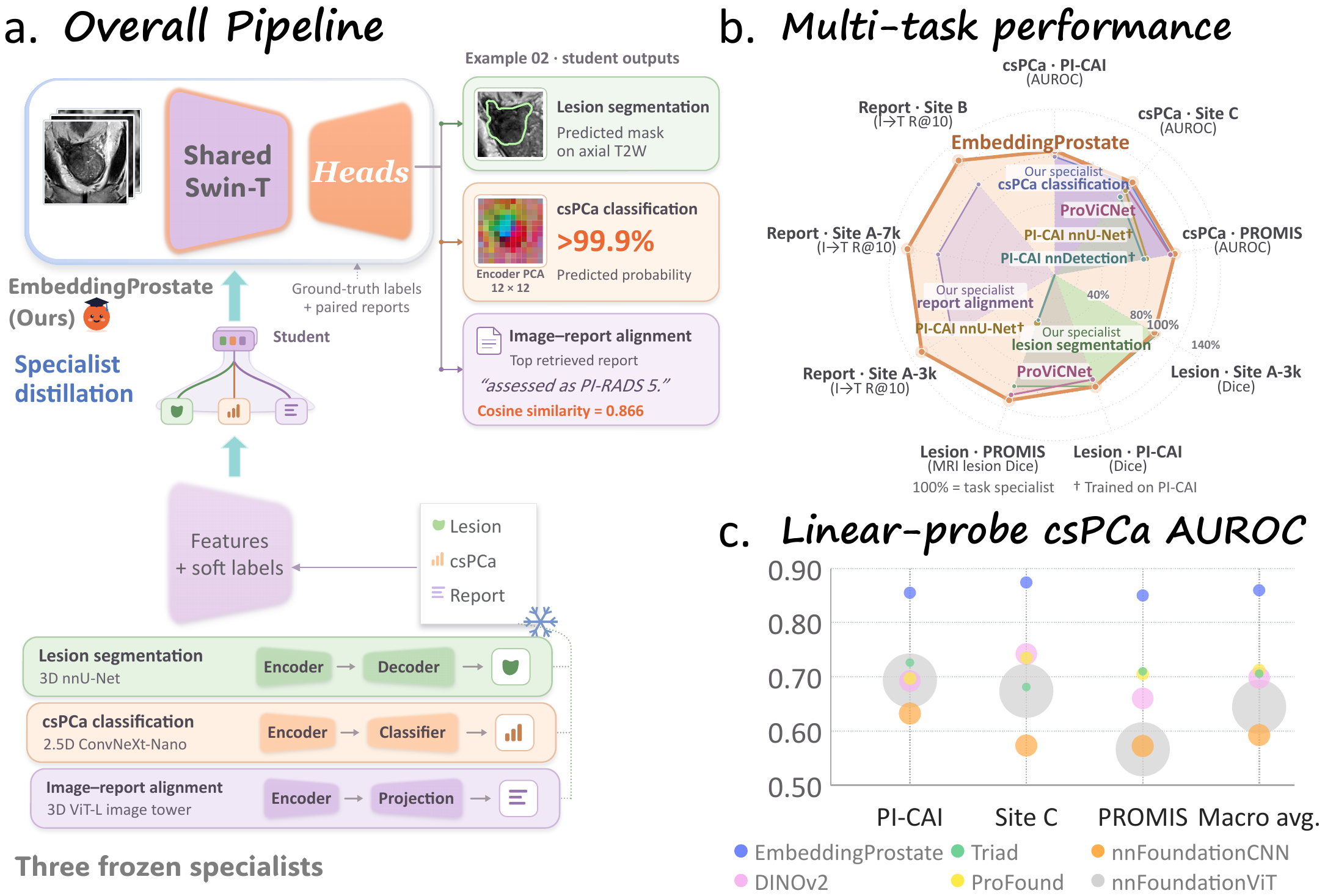}
  \caption{\textbf{EmbeddingProstate across tasks and cohorts.}
    \textbf{(a)} Student encoder distilled from three frozen specialists outlines a lesion, estimates the probability of clinically significant prostate cancer (csPCa), and retrieves a report for the same MRI examination.
    \textbf{(b)} Comparison of EmbeddingProstate with the baseline models and with each specialist model trained on the same training examples.
    \textbf{(c)} Linear-probe csPCa AUROC on PI-CAI, Site~C, and PROMIS with frozen encoders. Bubble area is proportional to the combined encoder and classifier parameter count.}
  \label{fig:embeddingprostate-overview}
\end{figure}

% Editable PowerPoint source: embeddingprostate_architecture.pptx.
\begin{figure*}[!t]
  \centering
  \includegraphics[width=\textwidth]{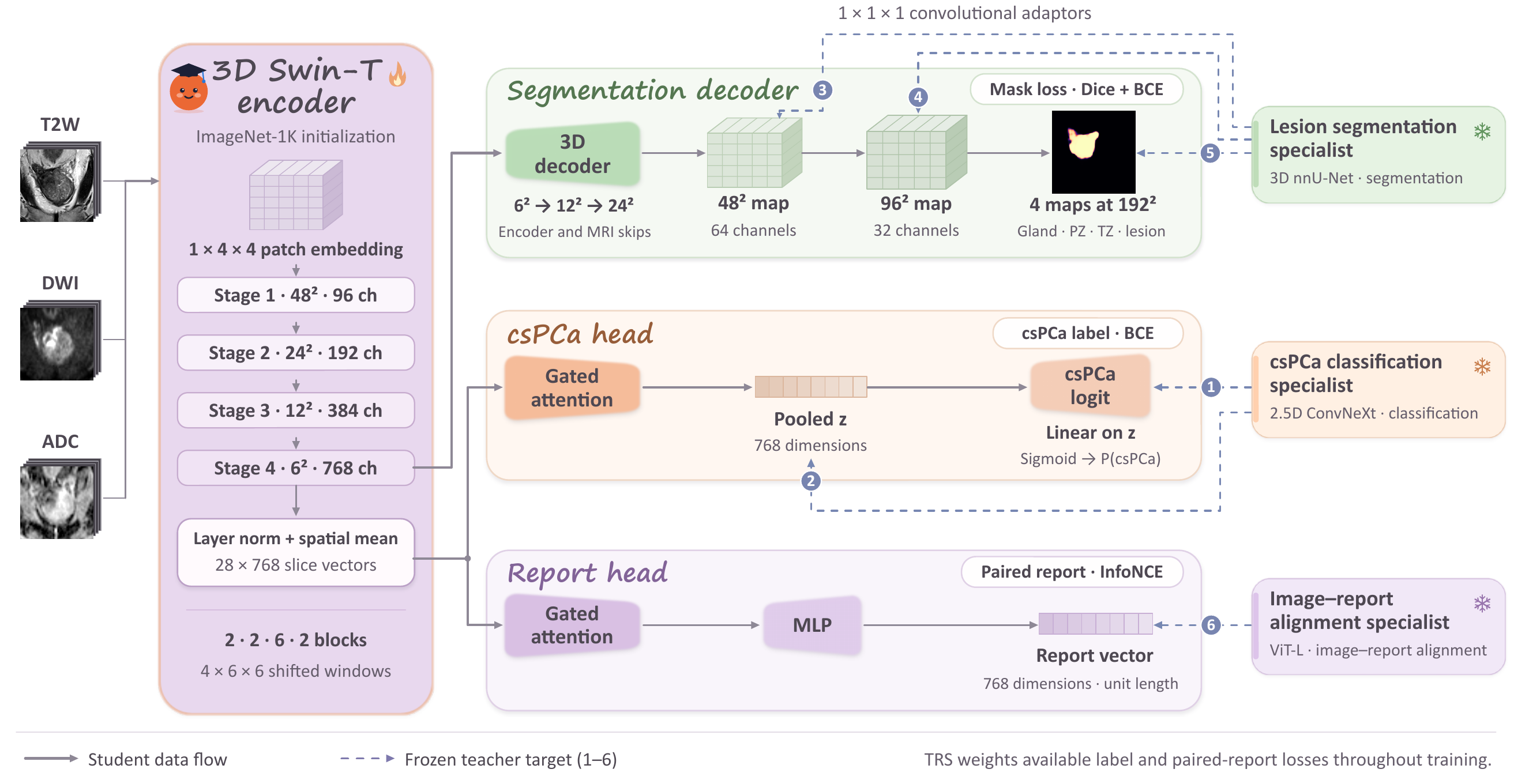}
  \caption{\textbf{Architecture and training of EmbeddingProstate.}
  A shared 3D Swin-T encoder takes T2W, DWI, and ADC volumes from one examination. In a single forward pass, a decoder segments the prostate gland, its zones, and lesions, while two heads pool features across slices to estimate the probability of clinically significant prostate cancer (csPCa) and produce an image embedding for report retrieval.
  At the numbered arrows, the student matches targets from three frozen specialists. These are the csPCa logit and pooled feature (1--2), decoder features and segmentation probability maps (3--5), and the image embedding (6).}
  \label{fig:embeddingprostate-architecture}
\end{figure*}

% --- P1: why prostate MRI matters; reader variability; shortage of expert readers
Prostate cancer is the second most frequently diagnosed cancer in men, with about 1.5 million new cases and 420,000 deaths in 2024~\cite{sung2026globocan}. The number of new cases per year is projected to double, from 1.4 million in 2020 to 2.9 million by 2040~\cite{james2024lancet}. Magnetic resonance imaging (MRI) is now recommended by the European Association of Urology guidelines for men with suspected disease~\cite{cornford2026eau}, as it detects more clinically significant prostate cancer (csPCa) than systematic biopsy and lets more than a quarter of men avoid a biopsy~\cite{ahmed2017promis,kasivisvanathan2018precision}; in the United States, the share of biopsies preceded by MRI rose from 0.5\% in 2007 to 35.5\% in 2022~\cite{soerensen2025trends}. Reading the examination, however, is difficult. A radiologist must combine T2-weighted, diffusion-weighted, and apparent diffusion coefficient (ADC) images into a five-point PI-RADS suspicion score~\cite{turkbey2019pirads}, and reader variability is high. Across 26 centers, only 35\% of the lesions scored PI-RADS 3 or higher proved to be csPCa, with wide variation between centers~\cite{westphalen2020variability}; among nine radiologists, the share of lesions scored 4 that proved to be csPCa ranged from 23\% to 65\%~\cite{sonn2019prostate}. By expert consensus, expert-level experience takes at least 1,000 examinations to acquire~\cite{derooij2020esur}, while demand for prostate MRI keeps rising and radiologists are in short supply~\cite{penzkofer2021esur,rcr2026census}. These findings show the need for accurate and scalable computer-aided interpretation of prostate MRI.

% --- P2: current AI models (three groups, the baselines of our tables) and their two limitations
Deep learning has become a good and reliable assistant. In the PI-CAI study, an AI system trained on 9,207 examinations detected csPCa with an area under the ROC curve (AUROC) of 0.91, against 0.86 for a pool of 62 radiologists~\cite{saha2024picai}. Current models fall into three groups, each built for a single purpose. \emph{Detection and segmentation models} localize csPCa. \emph{Vision--language models} align images with text. \emph{Self-supervised encoders} learn general-purpose features without labels. These models share two limitations. First, each serves a single purpose. A lesion segmentation model and a csPCa classifier do not describe what they find; a vision--language model returns text but no lesion map; and a pretrained self-supervised encoder performs no clinical task until a new head is trained on labeled target data. A radiologist's read, in contrast, answers three questions at once: whether csPCa is present, where it is, and how it should be described. Second, their supervision does not scale. A detection model can learn only from examinations with lesion annotations and a vision--language model only from those with text, so data labeled for one purpose is lost to the others. The most informative labels are also the scarcest: a csPCa label requires a biopsy and a lesion mask requires expert delineation, whereas a report is written for every examination.

% --- P3: the need for a unified model; why it is plausible; Platonic Representation Hypothesis; name the model
These limitations call for a \emph{unified} model for prostate MRI: one multi-task, multi-modal network whose single representation supports diagnosis, localization, and language, and which learns from every examination regardless of the label it carries. Such a model is plausible because the three tasks are closely related. A csPCa label, a lesion mask, and a radiology report describe the same gland: the label states that significant cancer is present, the mask states where it is, and the report describes its location, size, and appearance in words. Features that explain one of them should therefore help explain the others, so tasks trained together can learn from each other. This expectation agrees with the Platonic Representation Hypothesis~\cite{huh2024platonic}, which holds that networks trained with different objectives, on different data and modalities, converge toward a shared model of the reality that produced their data. The hypothesis also argues that each additional task constrains the representation, because fewer representations can solve many tasks than can solve one. A single encoder trained for diagnosis, localization, and language should therefore come closer to this shared representation than a model trained for one task, and our three specialists can be seen as partial estimates of it. We realize this idea in EmbeddingProstate (Figure~\ref{fig:embeddingprostate-overview}), a model trained on three clinical cohorts with three objectives that performs all three tasks with one encoder.

% --- P4: two challenges (no complete dataset; uneven overfitting of scattered labels)
Building it raised two challenges. First, no dataset provides all three kinds of supervision. Public prostate MRI datasets such as PI-CAI~\cite{saha2022picaidataset} and PROMIS~\cite{ahmed2017promis,wang2025promisderived} provide csPCa labels and lesion annotations but no radiology reports, and the image--report pairs behind recent MRI vision--language models are private~\cite{wang2026prostlm,yang2026deciphermr}. Hospital archives show the opposite pattern. In our three training cohorts, all 10,681 examinations have a report, but only 2,334 (22\%), all from one cohort, have csPCa labels and lesion masks. Supervision is thus scattered across cohorts and tasks. Second, scattered supervision is hard to balance. When the ground-truth labels of the three tasks are optimized together, they are fitted at very different rates: the csPCa labels are memorized, while image--report retrieval still benefits from its labels (Section~\ref{sec:method-trs}). No single loss weight or stopping point suits every task, and generic loss-balancing rules~\cite{kendall2018uncertainty,chen2018gradnorm} balance the scale or training rate of the losses without asking whether a label still has anything to teach.

% --- P5: our answer to both challenges; contributions
EmbeddingProstate addresses both challenges. We first train one specialist per task on the examinations that carry its labels, and then distill the three specialists into one student~\cite{hinton2015distilling,ranzinger2024amradio}. Because a trained specialist can be applied to any examination, every examination supervises every task, including the 78\% that have no csPCa label or lesion mask. Ground-truth labels are then added with \emph{Teacher-Residual Supervision} (TRS), which weights each label loss by the loss that the corresponding specialist still incurs on those labels, that is, by the information the labels add beyond what distillation has already transferred. Our contributions are summarized below.
\begin{itemize}
  \item We present EmbeddingProstate, a unified prostate MRI model whose single three-dimensional encoder supports patient-level csPCa classification, lesion segmentation, and image--report retrieval. To our knowledge, it is the first to combine these three tasks at the level of whole examinations and their radiology reports. To our knowledge, it will also be the first open-source image--report alignment model for prostate MRI.
  \item We show that supervision scattered across cohorts and tasks can be unified by distilling task-specific specialists. EmbeddingProstate learns from 10,681 examinations of three clinical cohorts, although only 22\% of them carry csPCa labels and lesion masks.
  \item We identify uneven overfitting when ground-truth labels are added to distillation and propose Teacher-Residual Supervision, which sets each label weight from its teacher's own loss and adds no tuned constant. It removes the choice of how much each label type should count, is the best label-stage variant or within seed variation of the best, and remains stable when training continues (Section~\ref{sec:method-trs}).
  \item On five prostate MRI datasets, including three external cohorts never used in training (PI-CAI, Site~C, and PROMIS), one EmbeddingProstate checkpoint matches its three specialists in csPCa classification and lesion segmentation, with average changes of +2.5\% in AUROC and +3.4\% in Dice, and improves image-to-report R@10 by 27.4\%. Its frozen encoder exceeds the best of five public encoders by 18.4\% in linear-probe csPCa AUROC on average.
\end{itemize}

\section{Related Work}
\label{sec:related}

% Naming rule for this section: a model is named only if it is a baseline in our tables; all other work is cited by number only.
% Exception (Eden, 2026-11-16): the four unified computer-vision models in the "Unified models" paragraph are named.
% --- P1: overview of prior work (classification; detection and segmentation; image-text; the shift to foundation models and VLMs). No limitations yet.
Deep learning models for prostate MRI have largely addressed a single task, mostly focusing on csPCa classification~\cite{yoo2019prostate,hamm2023interactive,cai2024fully,shao2025mriptpca} or on lesion detection and segmentation~\cite{schelb2019classification,cao2019focalnet,saha2021endtoend,duran2022prostattention,pellicer2022deep,bosma2023semisupervised} independently. More recent research has extended toward language, combining prostate MRI with radiology reports or PI-RADS knowledge through image--text learning~\cite{wang2026prostlm,ma2025medrationale,yuan2026kampnet,olmos2026mfmgeom}. Driven by the success of large-scale pretraining on natural images~\cite{radford2021clip,oquab2024dinov2}, medical image analysis increasingly focuses on foundation models~\cite{wang2026triad,harsy2026nnfoundation,wang2026profound} and vision--language models (VLMs)~\cite{zhang2025biomedclip,sellergren2026medgemma,khattak2024unimedclip,lu2025radclip} that are built to serve many tasks rather than one.

% --- P2: prostate specialist models (classification; detection and segmentation; image-report), each followed by its weakness
\noindent\textbf{Prostate specialist models.}
Specialist models have advanced each task of prostate MRI reading separately. For csPCa classification, deep networks predict a patient-level probability from the images~\cite{yoo2019prostate,cai2024fully,shao2025mriptpca} and have matched radiologists with patient-level labels alone~\cite{cai2024fully}. Others explain each lesion classification with PI-RADS features~\cite{hamm2023interactive}. Despite these advances, a patient-level classifier returns a probability without a lesion mask, and none of these models relates the examination to its report. For lesion detection and segmentation, the self-configuring nnU-Net~\cite{isensee2021nnunet} and nnDetection~\cite{baumgartner2021nndetection} serve as the public PI-CAI baselines~\cite{bosma2023semisupervised,saha2024picai}, and ProViCNet~\cite{lee2026provicnet} adapts a vision transformer pretrained on natural images through patch-level contrastive learning. Other models also grade the lesions they find~\cite{cao2019focalnet,duran2022prostattention,pellicer2022deep}. However, these models depend on voxel-level lesion annotations, which require expert delineation, and they produce no text. Prostate image--report models are the most recent. They pretrain an image encoder against clinical report text for diagnosis~\cite{wang2026prostlm}, train lesion segmentation together with textual rationales~\cite{ma2025medrationale}, or adapt general image--text models to prostate cancer classification~\cite{yuan2026kampnet,olmos2026mfmgeom}. However, in all of them the text supports diagnosis or segmentation, those trained on report text process two-dimensional slices rather than whole examinations~\cite{wang2026prostlm,ma2025medrationale}, and none reports image--report retrieval.

% --- P3: unified models (computer vision; medical imaging; prostate MRI has none; what the nearest methods lack; our answer)
\noindent\textbf{Unified models.}
In computer vision, recent work has moved toward unified models that perform many tasks within a single network. This trend is exemplified by models such as Uni-Perceiver~v2~\cite{li2023uniperceiverv2}, Florence-2~\cite{xiao2024florence2}, Uni3DL~\cite{li2024uni3dl}, and DINO-X~\cite{ren2024dinox}, which share one architecture across detection, segmentation, and captioning, and by models that obtain a single encoder by distilling several pretrained teachers~\cite{ranzinger2024amradio,wang2024samclip,sariyildiz2025dune}. In medical imaging, generalist models follow the same direction~\cite{zhao2025biomedparse,zhou2026medversa,blankemeier2026merlin}. However, in prostate MRI, unified models remain limited, and to our knowledge none performs csPCa classification, lesion segmentation, and image--report alignment within one network. Foundation models such as DINOv2~\cite{oquab2024dinov2}, Triad~\cite{wang2026triad}, nnFoundation~\cite{harsy2026nnfoundation}, and ProFound~\cite{wang2026profound} provide a pretrained encoder, but it must still be adapted separately to every task, with gains that vary from task to task~\cite{huang2026transfer}. VLMs such as BiomedCLIP~\cite{zhang2025biomedclip}, MedSigLIP~\cite{sellergren2026medgemma}, UniMed-CLIP~\cite{khattak2024unimedclip}, and RadCLIP~\cite{lu2025radclip} focus on image--text alignment and offer neither lesion segmentation nor csPCa classification. The closest unified efforts are a model for PSMA PET/CT that combines report generation, question answering, and lesion segmentation~\cite{xing2026psma}, and a general MRI model presented with qualitative results only~\cite{he2025omnimri}. To address the above-mentioned limitations of specialist, foundation, and vision--language models, our proposed EmbeddingProstate distills task specialists across cohorts with scattered labels, which enables a single encoder to learn csPCa classification, lesion segmentation, and image--report retrieval together.

\section{Method}
\label{sec:method}

% --- P1: overview (Eden's paragraph, grammar corrected)
The EmbeddingProstate model is multi-modal, as it reads across three MRI modalities: T2W, DWI, and ADC. Its multi-task ability is distilled from our three prostate specialist models (Figure~\ref{fig:embeddingprostate-architecture}), each optimized with its own objective and dataset for a different task. The backbone of the unified student model can be any deep learning model, such as a 2.5D ConvNeXt, a 3D U-Net, or a Transformer-based model, that has ImageNet~\cite{deng2009imagenet} initialization, because its strong transfer learning ability can be applied to prostate MRI tasks. A unified model can be hard to train on a highly uneven and imbalanced dataset, as it can overfit to the labels once it has already been optimized on the feature maps. Therefore, we introduce Teacher-Residual Supervision (TRS) to facilitate training and to control overfitting on the labels of some tasks.

\subsection{Specialist models}
\label{sec:method-specialists}
% The three bold run-in headings below are \noindent\textbf{...} instead of \paragraph{...}: no vertical space before them (Eden, 2026-11-16, to save space).
This study was conducted under institutional review board (IRB) oversight. For image preprocessing, we crop the image volume input to $96\times96\times84$~mm centered on the prostate to zoom in on the prostate gland area using a trained prostate gland segmentation model. The crop is resampled to $192\times192\times28$ voxels, and T2W, DWI, and ADC are its three channels. Only the image--report specialist reads the full field of view. Each modality is then percentile-normalized to $[-1,1]$, which puts examinations from different scanners on the same intensity range.

% --- P2
\noindent\textbf{csPCa classification specialist.}
For csPCa classification we use a 2.5D network built on a ConvNeXt-Nano~\cite{liu2022convnext} with ImageNet weights, to which we add two parts. T2W, DWI, and ADC each go through their own stem, and a small gate decides at every location how much to take from each of the three. The shared stages then encode the 28 slices one by one, and attention-based multiple-instance pooling~\cite{ilse2018attention} combines the slice features into one csPCa logit. We train with binary cross-entropy on the 2,334 labeled Site~A-3k training examinations. We trained eight seeds, and the student later copies their mean logit.

% --- P3
\noindent\textbf{Lesion segmentation specialist.}
We use a 3D full-resolution nnU-Net~\cite{isensee2021nnunet} with six stages and 30M parameters, and we keep its Dice plus cross-entropy loss and its augmentation. It predicts three masks: the peripheral zone, the transition zone, and the lesion. We train it for 500 epochs on the same 2,334 examinations, all of which have expert masks.

% --- P4
\noindent\textbf{Image--report alignment specialist.}
This specialist follows CLIP~\cite{radford2021clip}, with a 3D ViT-L for the examination and CXR-BERT~\cite{boecking2022cxrbert} for the report. We first pretrain the ViT-L on prostate MRI without labels, using DINOv3 self-supervision~\cite{simeoni2025dinov3}. Each sequence is encoded on its own, and the three are pooled into one 768-d embedding. For the image--report training we borrow the recipe of COLIPRI~\cite{wald2025colipri} and its three losses: the contrastive loss between an examination and its report, a loss that tells a report sentence from its negation, and a report-generation loss. We found that report sentence negation loss is very important for medical image radiologist reports that are very homogeneous, like prostate MRI. To these we add a term that aligns lesion phrases with image regions and a term that scores csPCa with text prompts on the labeled examinations. We train on all 10,681 examination--report pairs.

% --- P5: the student (lead-in, two bullets, training protocol)

\subsection{EmbeddingProstate student distillation}
\label{sec:method-setting}
The student $f_\theta$ is trained in two stages, and the three specialists stay frozen the whole time. We use all 10,681 training examinations from Site~A-3k, Site~A-7k, and Site~B. In Stage A the student only copies the specialists. We call this feature map supervision:
\begin{equation}
\mathcal{L}_{\mathrm{copy}}(\theta)=\frac{1}{|\mathcal{T}|}\sum_{T\in\mathcal{T}}\mathcal{L}_{T}(\theta).
\label{eq:copy}
\end{equation}
Here $\mathcal{L}_T$ is the loss for copying specialist $T$, both its outputs and its features. In Stage B we keep $\mathcal{L}_{\mathrm{copy}}$ and add the ground-truth (GT) labels, which is the topic of Section~\ref{sec:method-trs}.
\begin{itemize}
  \item \textbf{Architecture.} The student is a 3D Swin-T~\cite{liu2021swin,liu2022videoswin} that we inflate from ImageNet weights. Swin-T is the smaller of the two leading backbones on PI-CAI (Supplementary Table~\ref{tab:student-backbones}). The three sequences take the place of the three color channels. The patch size is $1\times4\times4$, which means that the 28 slices are never merged and each slice keeps its own tokens. csPCa uses attention-based multiple-instance pooling over the slices, segmentation uses a U-Net-style decoder that predicts the lesion and the two zones, and retrieval uses a projection to a 768-d embedding. The student has no text encoder of its own. Reports are embedded by the frozen text encoder of the image--report specialist.
  \item \textbf{Where the feature maps are copied.} From the csPCa specialist the student copies the mean logit and the pooled feature. From the segmentation specialist it copies the zone and lesion probability maps and the decoder feature maps at $48\times48$ and $96\times96$. From the image--report specialist it copies the 768-d image embedding. We use a KL divergence at temperature 2 for the logit, soft binary cross-entropy for the probability maps, cosine-based losses for the features, and a squared distance for the unit-norm embedding. Features go through a small adaptor, which we remove after training.
\end{itemize}
% Compact single-column version (2026-11-16, to save space): figures/figure3_training_dynamics_compact.pdf, a 2 x 2 grid drawn by
% figures/training_dynamics_source/plot_training_dynamics_compact.py from the same three-seed summaries.
% The previous full-width version is figures/figure1_training_dynamics.pdf (figure*, width=\textwidth).
\begin{figure}[t]
  \centering
  \includegraphics[width=\columnwidth]{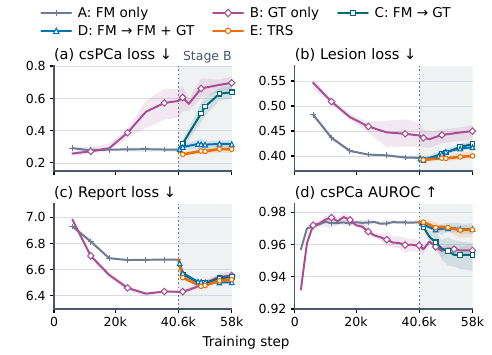}
  \caption{\textbf{Validation curves during training.} (a)--(c) Validation loss of each task and (d) validation csPCa AUROC for the variants of Table~\ref{tab:component-ablation} (FM: feature map supervision; GT: ground-truth labels). Lines are means over three seeds and bands show their range. Stage B is shaded and starts at step 40.6k. With uniform label weights (D), the csPCa and lesion losses rise in Stage B, whereas with TRS (E) they stay nearly flat.}
  \label{fig:training-dynamics}
\end{figure}

Training takes 58,000 steps with AdamW, a peak learning rate of $10^{-4}$, and a batch size of 8. The first 40,600 steps are Stage A and the last 17,400 are Stage B, at $0.3\times$ the learning rate. In Stage A we sample examinations uniformly. In Stage B each batch has 4 examinations from Site~A-3k, 2 from Site~A-7k, and 2 from Site~B, so every batch contains labeled examinations. We keep an exponential moving average of the weights and pick the checkpoint with the highest csPCa AUROC on the Site~A-3k validation set. The same checkpoint is used for all three tasks.

% --- P6: Teacher-Residual Supervision (the problem, then the function)
\subsection{Teacher-Residual Supervision}
\label{sec:method-trs}
Adding the labels in Stage B sounds simple, but this is where training went wrong for us. The three kinds of labels are not fitted at the same speed (three-seed means on Site~A-3k; Figure~\ref{fig:training-dynamics}). When a student is trained on the labels alone, its validation csPCa AUROC goes from 0.974 at step 12,000 down to 0.960 at step 40,600. By then its csPCa cross-entropy is 0.0001 on training and 0.584 on validation examinations, so the csPCa labels have simply been memorized. A student that only copies the specialists does not have this problem and stays at 0.973--0.974. But once we add the labels with equal weights after copying, the problem is back: the validation cross-entropy goes up from 0.282 to 0.317, and validation lesion Dice goes down from 0.625 to 0.601. We think the reason is simple. The csPCa and lesion specialists were trained on these same labels and fit them almost exactly. For most examinations, the label tells the student nothing that the specialist has not already told it. Reports are different. The image--report specialist still has a loss of 5.8 nats on the true pairs, so there the labels have much more left to teach.

TRS turns this observation into a rule: weight each label loss by what the label adds beyond its specialist. Write $p^k_\theta(\cdot\mid x)$ for the student's prediction for label type $k\in\mathcal{K}=\{\mathrm{cls},\mathrm{les},\mathrm{rep}\}$, and $q_k(\cdot\mid x)$ for the prediction of the frozen specialist. For a labeled example $(x,y)$, we call the loss that the specialist itself has on the label the \emph{teacher residual}:
\begin{equation}
s_k(x,y)=-\log q_k(y\mid x).
\label{eq:surprise}
\end{equation}
Note that $-\log p^k_\theta(y\mid x)=s_k(x,y)+\log\big(q_k(y\mid x)/p^k_\theta(y\mid x)\big)$. So the residual is exactly the part of the label loss that is left when the student matches its specialist. TRS uses the batch mean of this residual as the weight of the label loss,
\begin{equation}
w_k=\frac{1}{|\mathcal{M}_k|}\sum_{(x,y)\in\mathcal{M}_k}s_k(x,y),
\label{eq:w-trs}
\end{equation}
where $\mathcal{M}_k$ is the set of labeled examples of type $k$ in the batch. The Stage B objective is then
\begin{equation}
\mathcal{L}(\theta)=\mathcal{L}_{\mathrm{copy}}(\theta)
+\sum_{k\in\mathcal{K}}\lambda_k\,w_k\,\mathcal{L}^{\mathrm{lab}}_k(\theta),
\label{eq:objective}
\end{equation}
where $\mathcal{L}^{\mathrm{lab}}_k(\theta)=\mathbb{E}_{(x,y)}\big[-\log p^k_\theta(y\mid x)\big]$ is the label loss of type $k$, $\lambda_{\mathrm{cls}}=\lambda_{\mathrm{les}}=1$, and $\lambda_{\mathrm{rep}}=0.1$. For the label losses we use binary cross-entropy for csPCa, Dice plus binary cross-entropy for lesions, and InfoNCE between the student's embedding and the report embeddings of the frozen text encoder. Zone masks, where we have them, are a fourth label type and are weighted in the same way. The weights are computed from the frozen specialists. No gradient flows through them, and there is no constant to tune. In our runs the weights are about 0.04 for csPCa labels, 0.009 for lesion masks, and 5.8 for reports (Table~\ref{tab:design1-weights}). So the labels that a specialist already fits are learned slowly, and the labels that it fits loosely are learned at full strength.

\section{Results}
\label{sec:experiments}

% --- P1: what is compared and the note about one checkpoint
Classification is tested on three cohorts that none of our models has seen in training (PI-CAI, Site~C, and PROMIS), segmentation on PI-CAI and PROMIS, and retrieval on held-out patients of Site~B and Site~A-7k.

In the following experiments, each baseline and each specialist is a separate model trained for one task, whereas EmbeddingProstate uses one checkpoint for all tasks.

% Queue wide comparisons in reading order before their Results paragraphs.
% Task-specific panels preserve the original source-checkpoint scores.
% Native-system adaptation is in Supplementary Table S2, label tab:system-comparison-adaptation.
\begin{table*}[t]
  \caption{\textbf{Comparison with task-specific models.} (a) csPCa classification and (b) lesion segmentation on external cohorts, without any adaptation; (c) image-to-report retrieval on held-out patients. Cohort headers indicate whether our student and its corresponding specialist trained on data from that cohort. PI-CAI-trained baselines ($\dagger$) omit their PI-CAI scores. Values with $\pm$ are the mean and SD over three training runs. Bold marks the best result in each column.}
  \label{tab:system-comparison}
  \centering
  \footnotesize
  \setlength{\tabcolsep}{4pt}
  \renewcommand{\arraystretch}{1.0}
  \textbf{(a) csPCa classification (AUROC\EPup)}\par\vspace{2pt}
  \begin{tabularx}{\textwidth}{>{\raggedright\arraybackslash}p{.34\textwidth}*{3}{>{\centering\arraybackslash}X}}
    \toprule
    Model & PI-CAI~\cite{saha2022picaidataset}~\EPunseen & Site~C~\EPunseen & PROMIS~\cite{ahmed2017promis,wang2025promisderived}~\EPunseen \\
    \midrule
    \EPmodelrow \EPstudent{EmbeddingProstate} & 0.844$\pm$0.003 & \textbf{0.859}$\pm$0.001 & \textbf{0.830}$\pm$0.003 \\
    \EPmodelrow Our specialist: csPCa classification & 0.804$\pm$0.033 & 0.841$\pm$0.004 & 0.806$\pm$0.013 \\
    ProViCNet~\cite{lee2026provicnet} & \textbf{0.860} & 0.802 & 0.800 \\
    PI-CAI nnU-Net$^{\dagger}$~\cite{isensee2021nnunet} & \textemdash~\EPseen & 0.782 & 0.637 \\
    PI-CAI nnDetection$^{\dagger}$~\cite{baumgartner2021nndetection} & \textemdash~\EPseen & 0.725 & 0.616 \\
    \bottomrule
  \end{tabularx}
  \par\vspace{3pt}
  \textbf{(b) Lesion segmentation and localization}\par\vspace{2pt}
  \begin{tabularx}{\textwidth}{>{\raggedright\arraybackslash}p{.34\textwidth}*{4}{>{\centering\arraybackslash}X}}
    \toprule
    & \multicolumn{2}{c}{PI-CAI~\cite{saha2022picaidataset}~\EPunseen} & \multicolumn{2}{c}{PROMIS~\cite{ahmed2017promis,wang2025promisderived}~\EPunseen} \\
    \cmidrule(lr){2-3}\cmidrule(l){4-5}
    Model & Dice\EPup & Recall\EPup & Dice\EPup & Recall\EPup \\
    \midrule
    \EPmodelrow \EPstudent{EmbeddingProstate} & \textbf{0.467}$\pm$0.004 & \textbf{0.886}$\pm$0.004 & \textbf{0.236}$\pm$0.002 & \textbf{0.493}$\pm$0.020 \\
    \EPmodelrow Our specialist: lesion segmentation & 0.466$\pm$0.003 & 0.860$\pm$0.006 & 0.210$\pm$0.004 & 0.430$\pm$0.014 \\
    ProViCNet~\cite{lee2026provicnet} & 0.437 & 0.878 & 0.226 & 0.470 \\
    PI-CAI nnU-Net$^{\dagger}$~\cite{isensee2021nnunet} & \multicolumn{2}{c}{\textemdash~\EPseen} & 0.090 & 0.447 \\
    PI-CAI nnDetection$^{\dagger}$~\cite{baumgartner2021nndetection} & \multicolumn{2}{c}{\textemdash~\EPseen} & 0.085 & 0.393 \\
    \bottomrule
  \end{tabularx}
  \par\vspace{3pt}
  \textbf{(c) Image-to-report retrieval}\par\vspace{2pt}
  \begin{tabularx}{\textwidth}{>{\raggedright\arraybackslash}p{.34\textwidth}*{4}{>{\centering\arraybackslash}X}}
    \toprule
    & \multicolumn{2}{c}{Site~B~\EPseen} & \multicolumn{2}{c}{Site~A-7k~\EPseen} \\
    \cmidrule(lr){2-3}\cmidrule(l){4-5}
    Model & R@1\EPup & R@10\EPup & R@1\EPup & R@10\EPup \\
    \midrule
    \EPmodelrow \EPstudent{EmbeddingProstate} & 0.063$\pm$0.006 & \textbf{0.402}$\pm$0.007 & \textbf{0.091}$\pm$0.006 & \textbf{0.400}$\pm$0.014  \\
    \EPmodelrow Our specialist: image--report retrieval & 0.039 & 0.318 & 0.070 & 0.317  \\
    \midrule
    \multicolumn{5}{@{}l}{\textit{Released image--text checkpoints (no source fine-tuning in this study)}} \\
    BiomedCLIP~\cite{zhang2025biomedclip} & 0.000 & 0.031 & 0.002 & 0.022  \\
    MedSigLIP~\cite{sellergren2026medgemma} & 0.002 & 0.034 & 0.002 & 0.016  \\
    RadCLIP~\cite{lu2025radclip} & 0.002 & 0.027 & 0.003 & 0.019  \\
    UniMed-CLIP~\cite{khattak2024unimedclip} & 0.002 & 0.043 & 0.002 & 0.024  \\
    \midrule
    \multicolumn{5}{@{}l}{\textit{Full-model fine-tuning on source image--report pairs}~\EPseen} \\
    BiomedCLIP~\cite{zhang2025biomedclip} & \textbf{0.081}$\pm$0.014 & 0.378$\pm$0.005 & 0.084$\pm$0.002 & 0.381$\pm$0.002  \\
    MedSigLIP~\cite{sellergren2026medgemma} & 0.036$\pm$0.011 & 0.254$\pm$0.019 & 0.041$\pm$0.006 & 0.264$\pm$0.006  \\
    RadCLIP~\cite{lu2025radclip} & 0.027$\pm$0.007 & 0.249$\pm$0.013 & 0.058$\pm$0.008 & 0.307$\pm$0.018  \\
    UniMed-CLIP~\cite{khattak2024unimedclip} & 0.074$\pm$0.008 & 0.350$\pm$0.020 & 0.083$\pm$0.003 & 0.399$\pm$0.005  \\
    \bottomrule
  \end{tabularx}
  \par\vspace{2pt}
  {\footnotesize PI-CAI AUROC is within-center. Lesion recall is the share of reference lesions that a predicted region overlaps by at least 10\% of either region.}
\end{table*}

\begin{table*}[t]
  \caption{\textbf{Foundation model comparison.} Comparison of EmbeddingProstate's ability as a foundation model under frozen linear probing and full fine-tuning. $u$ marks known or possible prior overlap with the target cohorts, including the public PI-CAI studies used to pretrain ProFound. Values with $\pm$ are the mean and SD over three runs. Bold values identify EmbeddingProstate. Underlining marks the best result in each column.}
  \label{tab:representation-transfer}
  \centering
  \footnotesize
  \setlength{\tabcolsep}{3pt}
  \renewcommand{\arraystretch}{1.0}
  \begin{tabularx}{\textwidth}{>{\raggedright\arraybackslash}Xcccccc}
    \toprule
    csPCa AUROC\EPup & \multicolumn{2}{c}{PI-CAI~\cite{saha2022picaidataset}~\EPseen} & \multicolumn{2}{c}{Site~C~\EPseen} & \multicolumn{2}{c}{PROMIS~\cite{ahmed2017promis,wang2025promisderived}~\EPseen} \\
    \cmidrule(lr){2-3}\cmidrule(lr){4-5}\cmidrule(l){6-7}
    Encoder initialization & \shortstack{Frozen\\linear probe} & \shortstack{Full\\fine-tuning} & \shortstack{Frozen\\linear probe} & \shortstack{Full\\fine-tuning} & \shortstack{Frozen\\linear probe} & \shortstack{Full\\fine-tuning} \\
    \midrule
    DINOv2~\cite{oquab2024dinov2} & 0.692 & 0.652$\pm$0.029 & 0.742 & 0.711$\pm$0.117 & 0.660 & 0.667$\pm$0.069 \\
    Triad$^{u}$~\cite{wang2026triad} & 0.726 & 0.851$\pm$0.004 & 0.681 & 0.700$\pm$0.018 & 0.710 & 0.719$\pm$0.013 \\
    ProFound$^{u}$~\cite{wang2026profound} & 0.697 & 0.715$\pm$0.106 & 0.735 & 0.677$\pm$0.013 & 0.705 & 0.687$\pm$0.050 \\
    nnFoundationCNN$^{u}$~\cite{harsy2026nnfoundation} & 0.632 & 0.741$\pm$0.084 & 0.573 & 0.606$\pm$0.054 & 0.572 & 0.694$\pm$0.028 \\
    nnFoundationViT$^{u}$~\cite{harsy2026nnfoundation} & 0.693 & 0.820$\pm$0.004 & 0.674 & 0.743$\pm$0.023 & 0.566 & 0.653$\pm$0.009 \\
    \EPmodelrow \EPstudent{EmbeddingProstate encoder} & \EPresult{\EPbest{0.855}$\pm$0.001} & \EPresult{\EPbest{0.867}$\pm$0.014} & \EPresult{\EPbest{0.874}$\pm$0.009} & \EPresult{\EPbest{0.867}$\pm$0.011} & \EPresult{\EPbest{0.850}$\pm$0.005} & \EPresult{\EPbest{0.874}$\pm$0.005} \\
    \midrule
    \EPmodelrow Our specialist: csPCa classification & 0.851 & 0.855$\pm$0.005 & 0.834 & 0.827$\pm$0.045 & 0.849 & 0.856$\pm$0.010 \\
    \EPmodelrow Our specialist: lesion segmentation & 0.793 & 0.848$\pm$0.010 & 0.765 & 0.815$\pm$0.033 & 0.703 & 0.695$\pm$0.010 \\
    \EPmodelrow Our specialist: image--report retrieval & 0.787 & 0.837$\pm$0.006 & 0.744 & 0.801$\pm$0.022 & 0.708 & 0.738$\pm$0.024 \\
    \bottomrule
  \end{tabularx}
  \par\vspace{2pt}
  {\footnotesize PI-CAI AUROC is within-center. Protocols and data splits are given in Supplementary Section~\ref{sec:supplementary-details}.}
\end{table*}

\begin{figure*}[t]
  \centering
  % To change the size of the figure, edit only "\textwidth" below.
  % The citation number after "ProViCNet" in the column title is printed by LaTeX on top of the image, so it always matches the
  % reference list, and it is scaled together with the image. The lengths 242.78bp and 131.72bp belong to this PDF (495 x 251.4 bp; bp, not pt, because the PDF is measured in 1/72 inch).
  \resizebox{\textwidth}{!}{%
    \includegraphics{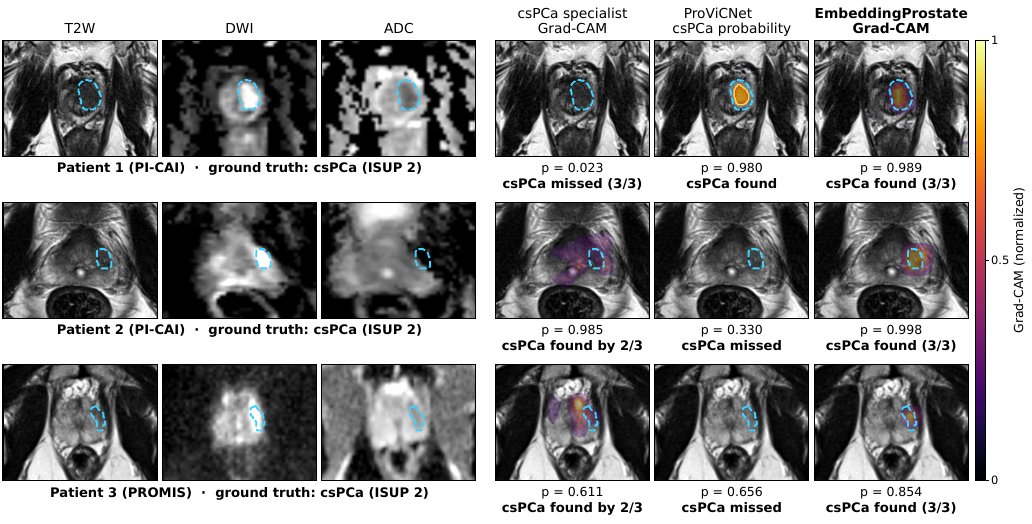}%
    \llap{\raisebox{242.78bp}[0pt][0pt]{\makebox[131.72bp][l]{\fontsize{6.6}{7}\selectfont\sffamily\cite{lee2026provicnet}}}}%
  }
  \caption{\textbf{Visualization of csPCa classification results on PI-CAI and PROMIS.} The input sequences (T2W, DWI, and ADC) are on the left, and the Grad-CAM~\cite{selvaraju2017gradcam} maps of our csPCa specialist and EmbeddingProstate and the csPCa probability map of ProViCNet are on the right. Each row represents a distinct patient with csPCa, and the csPCa result of each of our models is given over its three seeds. ProViCNet has only one trained model, so it has only one csPCa result.}
  \label{fig:gradcam-cases}
\end{figure*}

% --- P2
\noindent\textbf{csPCa classification results.}
We compare csPCa classification on the three external cohorts with ProViCNet, the PI-CAI baselines, and our csPCa specialist, as shown in Table~\ref{tab:system-comparison}(a). The results demonstrate that our unified model achieves superior or comparable performance to the task-specific models. EmbeddingProstate has the highest AUROC on Site~C and PROMIS, and it exceeds its own specialist on all three cohorts, so learning segmentation and retrieval in the same encoder does not reduce its diagnostic accuracy. ProViCNet is higher only on PI-CAI, and its accuracy drops on the other two cohorts, whereas EmbeddingProstate stays at a similar level on all three. The PI-CAI baselines were trained on PI-CAI itself and are therefore scored only on Site~C and PROMIS, where they remain well below EmbeddingProstate. Figure~\ref{fig:gradcam-cases} shows three patients with csPCa from PI-CAI and PROMIS that EmbeddingProstate finds, one that our csPCa specialist misses and two that ProViCNet misses, with its Grad-CAM on the reference lesion.

% csPCa Grad-CAM on three patients (Eden, 2026-11-16). Patients 2 and 3 = cases H3 (PI-CAI) and H28 (PROMIS, slice 13) of the candidate page, chosen by Eden;
% patient 1 = case H2 of the candidate page (slice 13), chosen by Eden: the csPCa specialist misses it. No badges; panels show the central 72 mm in height.
% 2026-11-17: vertical color bar for the Grad-CAM on the right; citation number after "ProViCNet" in the column title.
% The image is drawn on the server by UnifiedKD/paper/fig_gradcam_paper.py (first copy of each model, HiResCAM form of Grad-CAM,
% probabilities and 90%-specificity thresholds of the Table 1 scores). Notes on the figure: sec/X_suppl.tex, "Figures of the main paper".

% Empty metric cells are intentional. Populate only from matched, verified runs.
% Single-column table (2026-11-16, to save space): the SD is on a second line under each mean; row names are shortened
% (FM = the teacher's feature map supervision, defined in the caption); scaled to the column width.
\begin{table}[t]
  \caption{\textbf{Ablation of the training.} Rows A--E use the same student. All models are evaluated on held-out patients from their training cohort. TRS denotes Teacher-Residual Supervision, FM the teacher's feature map supervision, and GT ground-truth labels and paired reports; I$\to$T and T$\to$I denote image-to-text and text-to-image retrieval. Rows A--E are the mean and SD over three training seeds. Bold marks the highest value in each column. Underlining marks the best mean among C--E, which differ only in how the GT labels are added in Stage B.}
  \label{tab:component-ablation}
  \centering
  \footnotesize
  \setlength{\tabcolsep}{2pt}
  \renewcommand{\arraystretch}{0.85}
  \newcommand{\epsd}[1]{{\scriptsize$\pm$#1}}
  \resizebox{\columnwidth}{!}{%
  \begin{tabular}{lccccccc}
    \toprule
    & \multicolumn{7}{c}{Site~A-3k \EPseen} \\
    \cmidrule(l){2-8}
    & csPCa & \multicolumn{2}{c}{Lesion} & \multicolumn{2}{c}{I$\to$T} & \multicolumn{2}{c}{T$\to$I} \\
    \cmidrule(lr){2-2}\cmidrule(lr){3-4}\cmidrule(lr){5-6}\cmidrule(l){7-8}
    Method & AUROC\EPup & Dice\EPup & Recall\EPup & R@1\EPup & R@10\EPup & R@1\EPup & R@10\EPup \\
    \midrule
    A: FM only & 0.951 & 0.617 & \textbf{0.853} & 0.182 & 0.590 & 0.178 & 0.593 \\
     & \epsd{0.002} & \epsd{0.002} & \epsd{0.004} & \epsd{0.018} & \epsd{0.012} & \epsd{0.007} & \epsd{0.012} \\[2.4pt]
    B: GT only & 0.947 & 0.565 & 0.830 & 0.135 & 0.514 & 0.123 & 0.504 \\
     & \epsd{0.004} & \epsd{0.013} & \epsd{0.010} & \epsd{0.016} & \epsd{0.018} & \epsd{0.011} & \epsd{0.021} \\[2.4pt]
    C: FM $\to$ GT & 0.949 & \textbf{\EPbest{0.621}} & 0.839 & 0.219 & 0.647 & 0.191 & 0.635 \\
     & \epsd{0.003} & \epsd{0.005} & \epsd{0.007} & \epsd{0.013} & \epsd{0.011} & \epsd{0.010} & \epsd{0.005} \\[2.4pt]
    D: FM $\to$ FM + GT & 0.951 & 0.620 & 0.841 & 0.198 & 0.630 & 0.184 & 0.629 \\
     & \epsd{0.002} & \epsd{0.004} & \epsd{0.004} & \epsd{0.005} & \epsd{0.007} & \epsd{0.018} & \epsd{0.003} \\[2.4pt]
    \EPmodelrow E: TRS & \textbf{\EPbest{0.953}} & 0.619 & \EPbest{0.849} & \textbf{\EPbest{0.225}} & \textbf{\EPbest{0.651}} & \textbf{\EPbest{0.196}} & \textbf{\EPbest{0.648}} \\
    \EPmodelrow {\scriptsize\EPstudent{EmbeddingProstate}} & \epsd{0.002} & \epsd{0.001} & \epsd{0.006} & \epsd{0.016} & \epsd{0.015} & \epsd{0.021} & \epsd{0.005} \\
    \bottomrule
  \end{tabular}}
  \par\vspace{2pt}
  \parbox{\columnwidth}{\footnotesize Row A is the checkpoint at the end of Stage A; rows B--E use the checkpoint with the highest validation csPCa AUROC.}
\end{table}

% --- P3
\noindent\textbf{Lesion segmentation results.}
To assess lesion segmentation, we compare EmbeddingProstate with the same models and with our lesion segmentation specialist on PI-CAI and PROMIS, as shown in Table~\ref{tab:system-comparison}(b). EmbeddingProstate has the highest Dice and the highest lesion recall on both cohorts. Its Dice is on par with the specialist on PI-CAI and higher on PROMIS, and it finds more of the reference lesions on both. Dice is lower for every model on PROMIS, possibly because PI-CAI annotates only lesions with confirmed csPCa, whereas PROMIS annotates any lesion that a radiologist sees.

% Statistics must be recomputed from the applied per-task batch scalars.
% Single-column table (2026-11-16, to save space): row names without the "Our specialist:" prefix, scaled to the column width.
\begin{table}[t]
  \caption{\textbf{Teacher-Residual Supervision weights.} Weight $w_k$ of each GT label loss, measured on 2,000 Stage B batches from the source training cohorts \EPseen{}. The effective coefficient is the mean weight multiplied by $\lambda_k$ of Eq.~\eqref{eq:objective}.}
  \label{tab:design1-weights}
  \centering
  \footnotesize
  \setlength{\tabcolsep}{3pt}
  \renewcommand{\arraystretch}{1.2}
  \resizebox{\columnwidth}{!}{%
  \begin{tabular}{lcccc}
    \toprule
    Specialist model & \shortstack{Mean\\$w$} & \shortstack{Median\\$w$} & \shortstack{IQR\\$w$} & \shortstack{Effective\\coefficient} \\
    \midrule
    csPCa classification & 0.0413 & 0.00223 & 0.00134--0.00694 & 0.0413 \\
    Lesion segmentation & 0.00919 & 0.00686 & 0.00366--0.0125 & 0.00919 \\
    Image--report retrieval & 5.80 & 5.78 & 5.43--6.12 & 0.580 \\
    \bottomrule
  \end{tabular}}
\end{table}

% --- P4
\noindent\textbf{Image--report retrieval results.}
The public image--text models are built for single 2D images, and none of them reads a three-sequence prostate examination directly. We therefore give them each slice of the crop as one color image, with T2W, DWI, and ADC as its three channels, and average the slice embeddings, without adding any layer. As released, all four models fail on this task, with retrieval close to chance. For a fair comparison, we fine-tune each of them, both encoders and with its own loss, on the same examination--report pairs that EmbeddingProstate is trained on. The results in Table~\ref{tab:system-comparison}(c) demonstrate that EmbeddingProstate still achieves the best or a comparable R@10 on both cohorts and improves on its own image--report specialist by about 26\%. Only the fine-tuned BiomedCLIP ranks the correct report first more often on Site~B.

% --- P5
\noindent\textbf{Foundation model comparison.}
We further evaluate EmbeddingProstate as a foundation model, and the results are shown in Table~\ref{tab:representation-transfer}. The task heads are removed, and a new csPCa classifier is trained on each target cohort, either as a linear probe on the frozen encoder or by full fine-tuning. The frozen EmbeddingProstate encoder outperforms all public encoders on the three cohorts, with an average gain of 18.4\% in linear-probe AUROC over the best of them. It is even higher than every public encoder after full fine-tuning, although some of these encoders may have seen PI-CAI or PROMIS images in pretraining. It also matches or exceeds the encoders of our three specialists, highlighting that a representation learned from three tasks transfers at least as well as one learned from a single task. Adaptation with the original task heads is reported in Supplementary Table~\ref{tab:system-comparison-adaptation}.

% --- P6: ablation (two paragraphs under one heading)
\noindent\textbf{Ablation studies.}
We compare five ways to train the student in Table~\ref{tab:component-ablation}: feature map supervision only (A), GT labels only (B), feature map supervision followed by GT labels (C), feature map supervision followed by both with uniform label weights (D), and TRS (E). Feature map supervision is the main ingredient: compared with training on GT labels alone, it improves lesion Dice from 0.565 to 0.617 and image-to-report R@10 from 0.514 to 0.590. Adding the labels afterwards mainly improves retrieval, and TRS adds them best, reaching an R@10 of 0.651 together with the highest csPCa AUROC and lesion recall of the three label-stage variants. Its margins over the other two variants are small but consistent: TRS is the best variant on nine of the eleven metrics in Supplementary Table~\ref{tab:label-stage-variants}, including csPCa AUROC on the external PI-CAI cohort, and it is within 0.002 of the best on the other two.

The larger benefit of TRS is stability. Its weights (Table~\ref{tab:design1-weights}) are close to zero for csPCa labels and lesion masks and large for reports, and Figure~\ref{fig:training-dynamics} shows the same pattern in the student: when the labels are added with uniform weights, the validation losses of csPCa classification and lesion segmentation rise, while the report loss continues to improve. TRS therefore keeps learning from the reports and takes little from the labels that would be memorized. When training continues to the end of the schedule, it loses only 0.007 csPCa AUROC, against 0.021 for variant C.

\section{Conclusion}
\label{sec:conclusion}

In this work, we present EmbeddingProstate, a unified foundation model for multi-modal and multi-task prostate MRI. EmbeddingProstate combines csPCa classification, lesion segmentation, and image-report alignment in a single architecture, and to our knowledge it is the first to combine these three tasks at the level of whole examinations and their radiology reports. By distilling task-specific specialists, EmbeddingProstate unifies supervision scattered across cohorts and tasks and learns from 10,681 examinations of three clinical cohorts. In addition, we introduce Teacher-Residual Supervision, which weights each ground-truth label by what it adds beyond its specialist. It reduces the uneven overfitting seen when ground-truth labels are added to distillation and keeps training stable. Trained on three clinical cohorts in which most examinations have no csPCa label or lesion mask, EmbeddingProstate matches or surpasses its specialists and state-of-the-art single-task models on five prostate MRI datasets, three of them never used in training. As a foundation model, its frozen encoder also outperforms five public encoders in csPCa classification.

{
    \small
    \bibliographystyle{ieeenat_fullname}
    \bibliography{main}
}

% Supplementary results follow the main manuscript with independent S numbering.
% Export these pages separately for the supplementary submission.
\clearpage
\setcounter{page}{1}
\renewcommand{\thepage}{S\arabic{page}}
\setcounter{section}{0}
\renewcommand{\thesection}{S\arabic{section}}
\renewcommand{\theHsection}{supp.\arabic{section}}
\setcounter{table}{0}
\renewcommand{\thetable}{S\arabic{table}}
\renewcommand{\theHtable}{supp.\arabic{table}}
\setcounter{equation}{0}
\renewcommand{\theequation}{S\arabic{equation}}
\renewcommand{\theHequation}{supp.\arabic{equation}}
\maketitlesupplementary
% Keeps reference [12] (HiResCAM) in the reference list, so that every citation number stays the same as in the submitted main paper.
\nocite{draelos2020hirescam}

\section{Overview}
\label{sec:supplementary-overview}
This supplementary material gives the details that the main paper refers to and adds results that did not fit in it. Section~\ref{sec:supplementary-details} describes the datasets: the cohorts used to train the csPCa classification, lesion segmentation, and image--report specialists and the student, the cohorts used for testing, and the data splits. Section~\ref{sec:supplementary-implementation} gives the implementation details needed to reproduce the training: the full training configuration and the loss formulation. Section~\ref{sec:supplementary-results} presents extended results for the experiments of the main paper: the label-stage variants (Table~\ref{tab:label-stage-variants}), adaptation with the original task heads (Table~\ref{tab:system-comparison-adaptation}), the feature maps distilled from the segmentation specialist (Table~\ref{tab:copy-placement}), and the student backbones (Table~\ref{tab:student-backbones}). Section~\ref{sec:supplementary-limitations} discusses limitations and future work.

% All six tables are input here, in the order S1 to S6 (the order fixes their numbers), so that they float to the tops of the pages near their text.
% Generated by make_tableS1.py from collect.py's cells.json; do not edit numbers by hand.
% Included through sec/X_suppl.tex as Supplementary Table S1.
\begin{table*}[t]
  \caption{Label-stage variants on all metrics. Values are the mean and SD over three training seeds. Bold and underlining mark the best mean in each row.}
  \label{tab:label-stage-variants}
  \centering
  \small
  \setlength{\tabcolsep}{3pt}
  \renewcommand{\arraystretch}{1.12}
  \begin{tabular}{@{}llcc>{\columncolor{EPModelBlue}[\tabcolsep][0pt]}c@{}}
    \toprule
    Cohort & Metric & C & D & \shortstack{\EPstudent{EmbeddingProstate}\\E (TRS)} \\
    \midrule
    Site~A-3k \EPseen & csPCa AUROC\EPup & 0.949$\pm$0.003 & 0.951$\pm$0.002 & \textbf{\EPbest{0.953}}$\pm$0.002 \\
     & lesion Dice\EPup & \textbf{\EPbest{0.621}}$\pm$0.005 & 0.620$\pm$0.004 & 0.619$\pm$0.001 \\
     & lesion recall\EPup & 0.839$\pm$0.007 & 0.841$\pm$0.004 & \textbf{\EPbest{0.849}}$\pm$0.006 \\
     & I$\to$T R@10\EPup & 0.647$\pm$0.011 & 0.630$\pm$0.007 & \textbf{\EPbest{0.651}}$\pm$0.015 \\
    Site~A-7k \EPseen & I$\to$T R@10\EPup & 0.392$\pm$0.005 & 0.383$\pm$0.005 & \textbf{\EPbest{0.400}}$\pm$0.014 \\
    Site~B \EPseen & csPCa AUROC\EPup{} (patient benchmark) & 0.761$\pm$0.003 & 0.760$\pm$0.003 & \textbf{\EPbest{0.764}}$\pm$0.003 \\
     & I$\to$T R@10\EPup & 0.390$\pm$0.010 & 0.383$\pm$0.005 & \textbf{\EPbest{0.402}}$\pm$0.007 \\
    PI-CAI~\cite{saha2022picaidataset}~\EPunseen & csPCa AUROC, within-center\EPup & 0.839$\pm$0.004 & 0.841$\pm$0.002 & \textbf{\EPbest{0.844}}$\pm$0.003 \\
     & csPCa AUROC, pooled\EPup & 0.814$\pm$0.002 & 0.817$\pm$0.001 & \textbf{\EPbest{0.821}}$\pm$0.002 \\
     & lesion Dice\EPup & \textbf{\EPbest{0.469}}$\pm$0.011 & 0.467$\pm$0.006 & 0.467$\pm$0.004 \\
     & lesion recall\EPup & 0.882$\pm$0.008 & 0.880$\pm$0.001 & \textbf{\EPbest{0.886}}$\pm$0.004 \\
    \bottomrule
  \end{tabular}
  \par\vspace{2pt}
  {\footnotesize C is feature map supervision followed by GT labels, D is feature map supervision followed by both with uniform label weights, and E is TRS. The Site~B patient benchmark has 1,025 examinations (609 csPCa positive) of patients who are not in the training set.}
\end{table*}

\begin{table}[t]
  \caption{Adaptation on PI-CAI with the original task heads. csPCa AUROC and lesion Dice on the PI-CAI test set before (direct transfer) and after fine-tuning. Bold and underlining mark the best result in each column.}
  \label{tab:system-comparison-adaptation}
  \label{tab:native-adaptation-supplement}
  \centering
  \footnotesize
  \setlength{\tabcolsep}{3pt}
  \renewcommand{\arraystretch}{1.08}
  \resizebox{\columnwidth}{!}{%
  \begin{tabular}{@{}lcccc@{}}
    \toprule
    & \multicolumn{2}{c}{csPCa AUROC\EPup} & \multicolumn{2}{c}{Lesion Dice\EPup} \\
    \cmidrule(lr){2-3}\cmidrule(l){4-5}
    Model & \shortstack{Direct transfer\\\EPunseen} & \shortstack{Fine-tuned\\\EPseen} & \shortstack{Direct transfer\\\EPunseen} & \shortstack{Fine-tuned\\\EPseen} \\
    \midrule
    \EPmodelrow \EPstudent{EmbeddingProstate (TRS)} & 0.856$\pm$0.002 & \textbf{\EPbest{0.865}}$\pm$0.008 & \textbf{\EPbest{0.486}}$\pm$0.002 & \textbf{\EPbest{0.539}}$\pm$0.006 \\
    \EPmodelrow Our specialist: csPCa classification & 0.811$\pm$0.044 & 0.847$\pm$0.010 & \textemdash & \textemdash \\
    \EPmodelrow Our specialist: lesion segmentation & \textemdash & \textemdash & 0.475$\pm$0.008 & 0.532$\pm$0.005 \\
    ProViCNet~\cite{lee2026provicnet} & \textbf{\EPbest{0.866}} & 0.858$\pm$0.010 & 0.434 & 0.251$\pm$0.064 \\
    \bottomrule
  \end{tabular}}
  \par\vspace{2pt}
  \parbox{\columnwidth}{\footnotesize Values with $\pm$ are the mean and SD over three models, or over three fine-tuning seeds for ProViCNet. A dash means that the model has no head for that task.}
\end{table}

% Supplementary Table S3: placement of the segmentation specialist's feature maps (six runs, one seed, end of Stage A).
\begin{table*}[t]
  \caption{Ablation of the feature maps distilled from the segmentation specialist. Six choices of the layers whose feature maps the student copies (student $\leftarrow$ specialist), each trained with one seed for Stage A only. $48^2$ denotes a feature map with an in-plane size of $48\times48$, and likewise for the other sizes.}
  \label{tab:copy-placement}
  \centering
  \small
  \setlength{\tabcolsep}{4pt}
  \renewcommand{\arraystretch}{1.22}
  \begin{tabularx}{\textwidth}{>{\raggedright\arraybackslash}Xccccccc}
    \toprule
    & \multicolumn{4}{c}{Site~A-3k \EPseen} & \multicolumn{3}{c}{PI-CAI~\cite{saha2022picaidataset}~\EPunseen} \\
    \cmidrule(lr){2-5}\cmidrule(lr){6-8}
    Distilled layers (student $\leftarrow$ specialist) & \shortstack{csPCa\\AUROC\EPup} & \shortstack{Lesion\\Dice\EPup} & \shortstack{Lesion\\recall\EPup} & \shortstack{I$\to$T\\R@10\EPup} & \shortstack{csPCa\\AUROC\EPup} & \shortstack{Lesion\\Dice\EPup} & \shortstack{Lesion\\recall\EPup} \\
    \midrule
    Probability maps only (no feature map) & 0.952 & 0.614 & 0.865 & 0.591 & 0.843 & 0.468 & 0.892 \\
    \EPmodelrow \shortstack[l]{\EPstudent{EmbeddingProstate} (default, Stage A)\\Decoder $48^2$, $96^2$ $\leftarrow$ decoder $48^2$, $96^2$} & 0.951 & 0.618 & 0.854 & 0.591 & 0.845 & 0.466 & 0.885 \\
    Decoder $12^2$, $24^2$ $\leftarrow$ decoder $12^2$, $24^2$ & 0.950 & 0.612 & 0.848 & 0.570 & 0.846 & 0.457 & 0.890 \\
    Decoder, four levels $\leftarrow$ decoder, four levels & 0.951 & 0.620 & 0.857 & 0.577 & 0.845 & 0.461 & 0.883 \\
    Encoder $48^2$ $\leftarrow$ decoder $48^2$ & 0.950 & 0.615 & 0.865 & 0.584 & 0.842 & 0.464 & 0.890 \\
    Decoder $48^2$, $96^2$ $\leftarrow$ encoder $48^2$, $96^2$ & 0.951 & 0.614 & 0.857 & 0.591 & 0.842 & 0.462 & 0.894 \\
    \bottomrule
  \end{tabularx}
  \par\vspace{2pt}
  {\footnotesize The copies of the specialists' outputs are the same in every row.}
\end{table*}

% Supplementary Table S4 since 2026-11-16: moved out of the main paper to save space (input in sec/X_suppl.tex).
% U-Net row = train_ukd --student f3d (author-confirmed 2026-11-06).
% The report documents ConvNeXt-Nano/Tiny and 3D Swin-T/S, not Swin-Nano.
% Single-column table (2026-11-16, to save space): the SD and the input type are on a second line under each row; scaled to the column width.
\begin{table}[t]
  \caption{Student backbones. Results of five students trained with the same recipe, on the Site~A-3k test set and on the external PI-CAI cohort, as the mean and SD over three seeds. Bold marks the two highest PI-CAI AUROCs.}
  \label{tab:student-backbones}
  \centering
  \footnotesize
  \setlength{\tabcolsep}{3pt}
  \renewcommand{\arraystretch}{1.1}
  \newcommand{\epsd}[1]{{\scriptsize$\pm$#1}}
  \resizebox{\columnwidth}{!}{%
  \begin{tabular}{lcccccc}
    \toprule
    & & \multicolumn{4}{c}{Site~A-3k \EPseen} & \shortstack{PI-CAI~\cite{saha2022picaidataset}\\\EPunseen} \\
    \cmidrule(lr){3-6}\cmidrule(l){7-7}
    \shortstack[l]{Student\\{\scriptsize input}} & \shortstack{Params\\(M)} & \shortstack{csPCa\\AUROC\EPup} & \shortstack{Lesion\\Dice\EPup} & \shortstack{I$\to$T\\R@1\EPup} & \shortstack{I$\to$T\\R@10\EPup} & \shortstack{csPCa\\AUROC\EPup} \\
    \midrule
    U-Net~\cite{ronneberger2015unet} & 20.8 & 0.955 & 0.621 & 0.241 & 0.675 & 0.836 \\[-2.5pt]
    {\scriptsize 3D} & & \epsd{0.001} & \epsd{0.005} & \epsd{0.010} & \epsd{0.014} & \epsd{0.003} \\
    ConvNeXt-Nano~\cite{liu2022convnext,rw2019timm} & 18.3 & 0.956 & 0.561 & 0.226 & 0.652 & 0.834 \\[-2.5pt]
    {\scriptsize 2.5D} & & \epsd{0.003} & \epsd{0.002} & \epsd{0.009} & \epsd{0.027} & \epsd{0.005} \\
    ConvNeXt-Tiny~\cite{liu2022convnext} & 31.6 & 0.959 & 0.557 & 0.215 & 0.660 & 0.832 \\[-2.5pt]
    {\scriptsize 2.5D} & & \epsd{0.002} & \epsd{0.007} & \epsd{0.019} & \epsd{0.021} & \epsd{0.004} \\
    \EPmodelrow \EPstudent{EmbeddingProstate} & 33.0 & 0.953 & 0.619 & 0.225 & 0.651 & \EPresult{0.844} \\[-2.5pt]
    \EPmodelrow {\scriptsize Swin-T (Tiny)~\cite{liu2022videoswin}, 3D} & & \epsd{0.002} & \epsd{0.001} & \epsd{0.016} & \epsd{0.015} & \EPresult{\epsd{0.003}} \\
    \EPstudent{Swin-S (Small)}~\cite{liu2022videoswin} & 54.4 & 0.952 & 0.617 & 0.226 & 0.646 & \EPresult{0.847} \\[-2.5pt]
    {\scriptsize 3D} & & \epsd{0.003} & \epsd{0.003} & \epsd{0.014} & \epsd{0.023} & \EPresult{\epsd{0.001}} \\
    \bottomrule
  \end{tabular}}
  \par\vspace{2pt}
  \parbox{\columnwidth}{\footnotesize The U-Net student has a ConvNeXt-Nano encoder inflated to 3D, and the 2.5D ConvNeXt students use the architecture of our csPCa specialist.}
\end{table}

% Supplementary Table S5 (2026-11-24): cohorts, labels, splits and T2W image sizes.
% Counts and medians: UnifiedKD/paper/dataset_table.py and dataset_table_promis.py (CPU jobs 3231221, 3231224). Input after Table S4, so that it is numbered S5.
\begin{table*}[t]
  \caption{Datasets. The number of examinations in total and in each cohort, with the labels, the training, validation, and test split, and the T2W image size of each cohort.}
  \label{tab:datasets}
  \centering
  \footnotesize
  \setlength{\tabcolsep}{4pt}
  \renewcommand{\arraystretch}{1.12}
  \resizebox{\textwidth}{!}{%
  \begin{tabular}{@{}lllrrrrrcc@{}}
    \toprule
    Cohort & Source & Labels that we use & Examinations & csPCa positive & Train & Val & Test & \shortstack{Median T2W volume\\(voxels)} & \shortstack{Median spacing\\(mm)} \\
    \midrule
    \multicolumn{10}{@{}l}{\textit{Training cohorts}} \\
    Site~A-3k & internal & csPCa, zone and lesion masks, report & 2,912 & 1,396 (48\%) & 2,334 & 287 & 291 & $576\times576\times22$ & $0.38\times0.38\times3.3$ \\
    Site~A-7k & internal & report & 6,256 & \textemdash & 5,006 & 625 & 625 & $576\times576\times22$ & $0.38\times0.38\times3.3$ \\
    Site~B & internal & report & 4,193 & \textemdash & 3,341 & 437 & 415 & $512\times512\times27$ & $0.35\times0.35\times3.0$ \\
    Total & & & 13,361 & & 10,681 & 1,349 & 1,331 & & \\
    \midrule
    \multicolumn{10}{@{}l}{\textit{External test cohorts}} \\
    PI-CAI~\cite{saha2022picaidataset} & external & csPCa, lesion masks & 1,500 & 425 (28\%) & 900 & 299 & 301 & $384\times384\times21$ & $0.50\times0.50\times3.0$ \\
    Site~C & external & csPCa & 705 & 493 (70\%) & 493 & 70 & 142 & $512\times512\times23$ & $0.52\times0.52\times3.5$ \\
    PROMIS~\cite{ahmed2017promis,wang2025promisderived} & external & csPCa, lesion contours & 566 & 305 (54\%) & 340 & 113 & 113 & $512\times512\times26$ & $0.39\times0.39\times3.3$ \\
    \bottomrule
  \end{tabular}}
  \par\vspace{2pt}
  {\footnotesize A dash means that csPCa labels are not used for training. For the external cohorts, the split is used only when a model is adapted to the cohort; otherwise all examinations are tested.}
\end{table*}

% Supplementary Table S6 (2026-11-24): configuration of the two training stages of EmbeddingProstate.
% Values read from train_ukd.py, ukd_student_swin3d.py, the specialist config (lr, weight decay, layer decay, EMA, drop path) and the run log of <N>_igLabels_scalar_s0.
% Input after Table S5, so that it is numbered S6. Every row gives the value of both stages (Eden 2026-11-24).
\begin{table}[t]
  \caption{Training configuration of EmbeddingProstate. Full settings of the student and of its two training stages. The table lists the architecture, the data, the optimizer and schedule, and the loss weights of each stage.}
  \label{tab:training-configuration}
  \centering
  \footnotesize
  \setlength{\tabcolsep}{5pt}
  \setlength{\aboverulesep}{0pt}
  \setlength{\belowrulesep}{0pt}
  \renewcommand{\arraystretch}{1.22}
  \resizebox{\columnwidth}{!}{%
  \begin{tabular}{@{}l|c|c@{}}
    \toprule
    \multicolumn{1}{@{}l}{Parameter} & \multicolumn{1}{c}{Stage A} & Stage B \\
    \midrule
    \multicolumn{3}{@{}l}{\textit{Architecture}} \\
    \midrule
    Backbone & 3D Swin-T & 3D Swin-T \\
    Input (channels $\times$ slices $\times$ height $\times$ width) & $3\times28\times192\times192$ & $3\times28\times192\times192$ \\
    Patch size & $1\times4\times4$ & $1\times4\times4$ \\
    Window size & $4\times6\times6$ & $4\times6\times6$ \\
    Embedding dimension & 96 & 96 \\
    Depths & 2, 2, 6, 2 & 2, 2, 6, 2 \\
    Attention heads & 3, 6, 12, 24 & 3, 6, 12, 24 \\
    Stochastic depth & 0.1 & 0.1 \\
    Parameters & 33.0M & 33.0M \\
    Initialization & ImageNet-1K & Stage A weights (EMA) \\
    \midrule
    \multicolumn{3}{@{}l}{\textit{Training}} \\
    \midrule
    Training examinations & 10,681 & 10,074$^{*}$ \\
    Examinations with csPCa labels and lesion masks & not used & 2,188 \\
    Steps & 40,600 & 17,400 \\
    Batch size & 8 & 8 \\
    Examinations per batch (Site~A-3k / A-7k / B) & uniform sampling & 4 / 2 / 2 \\
    Optimizer & AdamW & AdamW \\
    AdamW $\beta_1$, $\beta_2$ & 0.9, 0.999 & 0.9, 0.999 \\
    Peak learning rate & $1\times10^{-4}$ & $3\times10^{-5}$ \\
    Learning-rate schedule & cosine & cosine \\
    Warm-up (share of the stage) & 5\% & 2\% \\
    Layer-wise learning-rate decay & 0.75 & 0.75 \\
    Weight decay & 0.05 & 0.05 \\
    Gradient clipping (norm) & 1.0 & 1.0 \\
    EMA decay of the weights & 0.998 & 0.998 \\
    Mixed precision & bfloat16 & bfloat16 \\
    \midrule
    \multicolumn{3}{@{}l}{\textit{Losses}} \\
    \midrule
    Objective & $\mathcal{L}_{\mathrm{copy}}$ & $\mathcal{L}_{\mathrm{copy}}$ + TRS-weighted label losses \\
    Weight of each copy loss & 1/3 & 1/3 \\
    $\lambda$ of csPCa labels / lesion masks / reports & not used & 1 / 1 / 0.1 \\
    Temperature of the logit copy & 2 & 2 \\
    Checkpoint & last step & highest validation csPCa AUROC \\
    Training seeds & 3 & 3 \\
    \bottomrule
  \end{tabular}}
  \par\vspace{2pt}
  \parbox{\columnwidth}{\footnotesize $^{*}$Stage B leaves out 607 examinations that were reserved for a variant we did not adopt.}
\end{table}

\section{Datasets}
\label{sec:supplementary-details}
We use six cohorts from three hospitals and two public datasets (Table~\ref{tab:datasets}): three clinical cohorts for training (Site~A-3k, Site~A-7k, and Site~B) and three external cohorts for testing (PI-CAI~\cite{saha2022picaidataset}, Site~C, and PROMIS~\cite{ahmed2017promis,wang2025promisderived}). Site~A-3k and Site~A-7k come from the same hospital, and PI-CAI and PROMIS are public.

\noindent\textbf{csPCa classification.}
The csPCa labels come from Site~A-3k, which has 2,912 examinations, 1,396 of them (48\%) csPCa positive. We train on 2,334 examinations (1,116 positive) and keep 287 for validation and 291 for testing. Its median T2W volume has $576\times576\times22$ voxels at $0.38\times0.38\times3.3$~mm.

\noindent\textbf{Lesion segmentation.}
The same Site~A-3k examinations carry expert masks of the peripheral zone, the transition zone, and the lesions, so the segmentation specialist uses the same split.

\noindent\textbf{Image--report alignment.}
Every examination of the three training cohorts has a radiology report. This gives 10,681 examination--report pairs for training: 2,334 from Site~A-3k, 5,006 from Site~A-7k, and 3,341 from Site~B. Site~A-7k and Site~B contribute reports only.

\noindent\textbf{External test cohorts.}
PI-CAI has 1,500 public examinations from three centers (425 csPCa positive), Site~C has 705 examinations from one hospital (493 positive), and PROMIS has 566 patients (305 positive), whose reference is a template biopsy with Grade Group $\geq$ 2 counted as positive. None of our models was trained on these three cohorts. For the foundation model comparison (main-paper Table~\ref{tab:representation-transfer}), each cohort is divided into the fixed training, validation, and test split of Table~\ref{tab:datasets}. Every encoder reads the same crop of T2W, DWI, and ADC and receives a new csPCa classifier, trained either as a linear probe on the frozen encoder or by full fine-tuning for 20 epochs, and the hyperparameters are chosen on the validation split before the test split is scored.

Together, these resources amount to more than 16,000 examinations, making our work one of the most extensive among open-source models for prostate MRI (to our knowledge).

\noindent\textbf{Evaluation details.}
For PI-CAI, which has three centers, csPCa AUROC is computed within each center and averaged over the centers, weighted by their number of positive--negative pairs. The PI-CAI lesion reference is the annotation released by its organizers, drawn by a human expert for 220 of the 425 csPCa-positive examinations and derived by the organizers' AI model for the other 205. The PI-CAI baselines ($\dagger$) and ProViCNet are the released checkpoints, and nnU-Net and nnDetection were trained on PI-CAI, so their PI-CAI scores are not reported. For retrieval, each test examination ranks all reports of its test set (415 for Site~B, 625 for Site~A-7k, and 291 for Site~A-3k), so chance for R@10 is about 0.02 to 0.03, and the image--text baselines are fine-tuned on the same 10,681 pairs as EmbeddingProstate. In main-paper Figure~\ref{fig:gradcam-cases}, a patient counts as found when the csPCa probability reaches the model's own threshold at 90\% specificity.

\section{Implementation details}
\label{sec:supplementary-implementation}
Table~\ref{tab:training-configuration} reports the full configuration of EmbeddingProstate: the architecture of the student and, for each of its two training stages, the data, the hyperparameters, and the losses. Together with the loss formulation below, it gives what is needed to reproduce the training.

\subsection{Loss formulation}
In Stage A the student is trained with feature map supervision only (main-paper Eq.~\eqref{eq:copy}), which is the mean of three copy losses, one for each specialist:
\begin{equation}
\mathcal{L}_{\mathrm{copy}}
=\tfrac{1}{3}\Big(
\underbrace{\mathcal{L}^{\mathrm{copy}}_{\mathrm{cls}}}_{\text{csPCa}}
+\underbrace{\mathcal{L}^{\mathrm{copy}}_{\mathrm{seg}}}_{\text{Segmentation}}
+\underbrace{\mathcal{L}^{\mathrm{copy}}_{\mathrm{rep}}}_{\text{Image--report}}
\Big).
\label{eq:supp-copy}
\end{equation}
In Stage B the label losses are added (main-paper Eq.~\eqref{eq:objective}):
\begin{equation}
\begin{aligned}
\mathcal{L}=\mathcal{L}_{\mathrm{copy}}
&+\underbrace{w_{\mathrm{cls}}\,\mathcal{L}^{\mathrm{lab}}_{\mathrm{cls}}}_{\text{csPCa labels}}
+\underbrace{w_{\mathrm{les}}\,\mathcal{L}^{\mathrm{lab}}_{\mathrm{les}}}_{\text{Lesion masks}}\\
&+\underbrace{0.1\,w_{\mathrm{rep}}\,\mathcal{L}^{\mathrm{lab}}_{\mathrm{rep}}}_{\text{Reports}},
\end{aligned}
\label{eq:supp-objective}
\end{equation}
where $w_k$ is the TRS weight of main-paper Eq.~\eqref{eq:w-trs} and:
\begin{align}
\mathcal{L}^{\mathrm{copy}}_{\mathrm{cls}} &= \tfrac{1}{2}\,\tau^{2}\,\mathrm{KL}_{\tau}\big(z^{T}\,\|\,z\big)+\tfrac{1}{2}\big(1-\cos(h,h^{T})\big)\\
\mathcal{L}^{\mathrm{copy}}_{\mathrm{seg}} &= \tfrac{1}{2}\big(\mathrm{BCE}(\hat{m},m^{T})+\mathrm{Dice}(\hat{m},m^{T})\big)+\tfrac{1}{2}\,\mathcal{L}_{\mathrm{map}}\\
\mathcal{L}^{\mathrm{copy}}_{\mathrm{rep}} &= \lVert e-e^{T}\rVert_{2}^{2}\\
\mathcal{L}^{\mathrm{lab}}_{\mathrm{cls}} &= \mathrm{BCE}(z,y)+0.3\,\mathrm{BCE}(z_{\mathrm{slice}},y)\\
\mathcal{L}^{\mathrm{lab}}_{\mathrm{les}} &= \mathrm{BCE}(\hat{m}_{\mathrm{les}},m)+\mathrm{Dice}(\hat{m}_{\mathrm{les}},m)\\
\mathcal{L}^{\mathrm{lab}}_{\mathrm{rep}} &= -\log\frac{\exp(\kappa\,e^{\top}t^{+})}{\sum_{j}\exp(\kappa\,e^{\top}t_{j})}
\end{align}
Here $z$, $h$, $\hat{m}$, and $e$ are the csPCa logit, the pooled feature, the predicted zone and lesion maps, and the unit-norm image embedding of the student, and the superscript $T$ marks the same output of the frozen specialist. $\mathrm{KL}_{\tau}$ is the Kullback--Leibler divergence between the two csPCa probabilities at temperature $\tau=2$. $\mathcal{L}_{\mathrm{map}}$ compares the decoder feature maps of the student and of the segmentation specialist at $48\times48$ and $96\times96$, with $0.9$ times one minus their cosine similarity plus $0.1$ times a smooth L1 distance, on voxels sampled from the lesion, the gland, and the background. In the label losses, $y$ is the csPCa label, $z_{\mathrm{slice}}$ are the slice-level logits of an auxiliary head, and $m$ is the lesion mask. $t^{+}$ is the embedding of the examination's own report and $t_{j}$ runs over the training reports, all embedded by the frozen text encoder, with $\kappa=15.06$, the logit scale of the image--report specialist. Reports of the same patient or with identical text are left out of the sum. Zone masks are a fourth label type with the same loss as the lesion masks and their own TRS weight.

The specialists themselves are trained as described in the main paper: binary cross-entropy for the csPCa specialist, Dice plus cross-entropy for the segmentation specialist, and the contrastive, negation, and report-generation losses with our two additional terms for the image--report specialist.

\section{Extended results}
\label{sec:supplementary-results}
This section presents extended results for the experiments discussed in the main paper. In the supplementary tables, blue shading marks our models, and the tags \textit{(seen)} and \textit{(unseen)} indicate whether a cohort was used for training, as in the main paper.

\noindent\textbf{Detailed results of the label-stage variants.}
In the main paper (Table~\ref{tab:component-ablation}), we compared the ways of adding the GT labels only on the Site~A-3k test set. Here in Table~\ref{tab:label-stage-variants}, we provide all eleven metrics that we scored for the three label-stage variants, on held-out patients of Site~A-3k, Site~A-7k, and Site~B and on the external PI-CAI cohort. We compare TRS (E) against feature map supervision followed by GT labels (C) and feature map supervision followed by both with uniform label weights (D). Across the eleven metrics, TRS achieves the best performance on nine and is within 0.002 of the best on the other two, which are the lesion Dice scores. It also has the highest csPCa AUROC on PI-CAI, a cohort that was not used in training, demonstrating that its small but consistent margin holds beyond the training cohorts.

\noindent\textbf{Adaptation with the original task heads.}
In the main paper (Table~\ref{tab:representation-transfer}), we evaluated adaptation to a new cohort only for csPCa classification, with the task heads removed and a new classifier trained on each encoder. Here in Table~\ref{tab:system-comparison-adaptation}, we provide the results of fine-tuning each complete model on PI-CAI with its original task heads, for both csPCa classification and lesion segmentation, before (direct transfer) and after fine-tuning. We compare EmbeddingProstate against its csPCa classification and lesion segmentation specialists and ProViCNet~\cite{lee2026provicnet}. The model for evaluation is selected on the validation set. After fine-tuning, EmbeddingProstate's mean csPCa AUROC increases from 0.856 to 0.865 and its mean lesion Dice from 0.486 to 0.539. Among the compared models, it achieves the highest mean lesion Dice both before and after fine-tuning, and the highest mean csPCa AUROC after fine-tuning. These results support adapting the unified model to both tasks on PI-CAI with its original task heads.

\noindent\textbf{Effect of the distilled feature maps.}
In the main paper (Section~\ref{sec:method-setting}), we stated only that the student copies the decoder feature maps of the segmentation specialist at $48\times48$ and $96\times96$, without comparing other choices. Here in Table~\ref{tab:copy-placement}, we provide an ablation of six placements of this copy, on the Site~A-3k test set and on the external PI-CAI cohort. We compare the default against copying only the probability maps (no feature map), deeper decoder maps, all four decoder levels, and two placements that use encoder maps. All six placements perform similarly on the three tasks, and no placement is consistently better than the default, demonstrating that EmbeddingProstate is robust to where the feature maps are copied. We therefore kept the decoder maps at $48\times48$ and $96\times96$, which combine the high-level features of the deep layers with the low-level detail of the skip connections.

\noindent\textbf{Effect of the student backbone.}
In the main paper (Section~\ref{sec:method}), we stated that the backbone of the student can be any deep learning model with ImageNet initialization and that Swin-T is the smaller of the two leading backbones on PI-CAI, but we reported results only for Swin-T. Here in Table~\ref{tab:student-backbones}, we provide the full comparison of five student backbones trained with the same recipe, on the Site~A-3k test set and on the external PI-CAI cohort. We compare the 3D Swin-T of EmbeddingProstate against a 3D U-Net, two 2.5D ConvNeXts (the architecture of our csPCa specialist, in two sizes), and a larger 3D Swin-S. All five students reach similar csPCa AUROC and retrieval on the internal test set, the 3D students segment lesions better than the 2.5D ones, and the two Swin students achieve the highest csPCa AUROC on PI-CAI. These results validate that our training recipe is not tied to one architecture and support the choice of Swin-T, the smaller of the two backbones that transfer best to external data.

\noindent\textbf{Effect of ImageNet initialization.}
In the main paper (Section~\ref{sec:method}), we required the backbone of the student to have ImageNet initialization. An ablation with the ConvNeXt-Nano student and one seed confirms this requirement: when the student starts from random weights, its csPCa AUROC at the end of Stage A falls from 0.958 to 0.930 and its image-to-report R@10 from 0.591 to 0.344.

\section{Limitations and future work}
\label{sec:supplementary-limitations}
Our work has a few limitations. First, because public prostate MRI datasets rarely include radiology reports, image--report retrieval is evaluated only on the test sets of our internal cohorts, and an evaluation on external cohorts with reports remains to be done. Second, EmbeddingProstate supports its three tasks with one shared representation, but the tasks run in parallel and not in sequence. In the future, we plan to develop a unified model that performs these tasks in sequence, as in the clinical procedure: it first localizes the lesion, then classifies csPCa, and finally generates the report with a language decoder on top of our encoder. We believe that a unified model trained in this way, by distillation or by another method, could show emergent reasoning from one step to the next and learn a more generalizable representation, for prostate MRI and for medical image analysis in general.

\section*{LLM Usage Statement}
We used Claude (Anthropic) to assist with writing and implementing code for this study, including multi-step implementation tasks.
We used Claude and ChatGPT (OpenAI) to polish the manuscript and correct grammar.
ChatGPT also helped us prepare the model architecture diagrams.
The authors take responsibility for the code, analyses, figures, and final manuscript.

\end{document}